# 3D Convolutional Encoder-Decoder Network for Low-Dose CT via Transfer Learning from a 2D Trained Network

Hongming Shan, Yi Zhang*, *Member, IEEE*, Qingsong Yang, Uwe Kruger, *Senior Member, IEEE*, Mannudeep K. Kalra, Ling Sun, Wenxiang Cong, Ge Wang*, *Fellow, IEEE*

*Abstract*—Low-dose computed tomography (CT) has attracted major attention in the medical imaging field, since CT-associated x-ray radiation carries health risks for patients. The reduction of the CT radiation dose, however, compromises the signal-to-noise ratio, which affects image quality and diagnostic performance. Recently, deep-learning-based algorithms have achieved promising results in low-dose CT denoising, especially convolutional neural network (CNN) and generative adversarial network (GAN) architectures. This article introduces a Conveying Path-based Convolutional Encoder-decoder (CPCE) network in 2D and 3D configurations within the GAN framework for low-dose CT denoising. A novel feature of this approach is that an initial 3D CPCE denoising model can be directly obtained by extending a trained 2D CNN, which is then fine-tuned to incorporate 3D spatial information from adjacent slices. Based on the transfer learning from 2D to 3D, the 3D network converges faster and achieves a better denoising performance when compared to a training from scratch. By comparing the CPCE network with recently published work based on the simulated Mayo dataset and the real MGH dataset, we demonstrate that the 3D CPCE denoising model has a better performance in that it suppresses image noise and preserves subtle structures.

## I. INTRODUCTION

Computed tomography (CT), utilizing x-ray radiation to create internal images of the body, is a widely used imaging modality in clinical, industrial, and other applications [1]. The widespread use of CT, however, has raised public concerns that, while CT helps a large number of patients, additional cancer cases could be induced by CT-related x-ray radiation [2], [3]. As the data from National Lung Screening Trial indicate, annual lung cancer screening for three years with low-dose CT (LDCT) resulted in 20% fewer lung cancer-related deaths [4]. Although CT must be used in this and other important tasks, decreasing the radiation dose as much as possible has, consequently, become a trend in CT-related research over the past decades. The reduction of radiation dose, however, increases noise and introduces artifacts in reconstructed images, which may adversely affect associated diagnostics based on LDCT images.

One way to address this is to reduce the image noise by filtering. This, however, is a challenging and ill-posed problem. Convolutional neural networks (CNNs) have recently shown their potential for image denoising [5]–[9]. Various denoising models based on CNNs have been proposed with different network architectures and objective functions for LDCT denoising, including 2D CNNs [5], [9], 3D CNN [7], residual encoder-decoder CNN [6], and cascaded CNN [8]. Different objective functions include the mean squared error (MSE) [5]–[8], the adversarial loss [7], [9], and the perceptual loss [9]. Different network architectures and objective functions, however, can have a profound impact upon the learning process: the architecture determines the complexity of the denoising model, while the loss function controls how to learn the denoising model from images and/or data.

In practice, a radiologist can extract pathological information more accurately and more reliably by looping through adjacent slices. Thus, there is a great opportunity to optimize networks and extend them from 2D to 3D so that the denoising model can recover more structural details. This is directly addressed in this article by introducing a novel 3D Conveying Path-based Convolutional Encoder-decoder (CPCE) network in the Generative Adversarial Network (GAN) framework. The main benefit of this architecture is to utilize the 3D spatial information from adjacent slices. More specifically, we first introduce a 2D CPCE as the denoising model, train this model in the GAN framework with both adversarial and perceptual losses, and then directly extend the trained 2D denoising model to 3D. In summary, the contributions of this paper are as follows:

- A 2D Conveying Path-based Convolutional Encoder-decoder (CPCE) network is introduced as the 2D denoising model;
- A 3D CPCE network is obtained by directly extending the above 2D model, which can utilize the 3D spatial information to enhance the 2D denoising performance;

Asterisk indicates corresponding author.

This work was supported in part by the National Natural Science Foundation of China under Grant 61671312, the Science and Technology Project of Sichuan Province of China under Grant 2018HH0070, and in part by the National Institute of Biomedical Imaging and Bioengineering/National Institutes of Health under Grant R01 EB016977 and Grant U01 EB017140.

H. Shan, Q. Yang, U. Kruger, W. Cong and G. Wang* are with Department of Biomedical Engineering, Rensselaer Polytechnic Institute, Troy, NY, 12180 USA (e-mail:{shanh, yangq4, krugeu, congw, wangg6}@rpi.edu).

Y. Zhang* is with the College of Computer Science, Sichuan University, Chengdu, 610065 China (e-mail: yzhang@scu.edu.cn).

M. K. Kalra is with Department of Radiology, Massachusetts General Hospital, Harvard Medical School, Boston, MA, 02114 USA (e-mail: mkalra@mgh.harvard.edu).

L. Sun is with Huaxi MR Research Center (HMRRC), Department of Radiology, West China Hospital, Sichuan University, Chengdu, 610041 China (e-mail: 251834489@qq.com).

IEEE TRANSACTIONS ON MEDICAL IMAGING 2and
- A simple yet effective transfer learning strategy is introduced to initialize the weights of a 3D CPCE network from a trained 2D model. Such an initialized 3D network starts with the same denoising performance of the trained 2D network and yields improvements relative to the established benchmark.

The paper is organized as follows. Section II surveys the noise reduction methods for LDCT and the GAN framework. Section III introduces the 2D CPCE as the denoising model and its 3D version, and explains how to extend a trained 2D denoising model to a 3D counterpart for learning additional spatial information. This is followed by comprehensively comparing the introduced CPCE models with recently-published competitive methods in Section IV. Finally, Section V presents a concluding summary of this work.

## II. RELATED WORK

### A. Noise reduction for LDCT

Noise reduction algorithms for LDCT can be categorized into (i) sinogram filtration [10]–[12], (ii) iterative reconstruction [13]–[19], and (iii) post-processing technique [5]–[9].

Sinogram filtration-based techniques perform on either raw data or log-transformed data before image reconstruction such as filtered backprojection (FBP). In the data domain, the noise characteristic is well known, allowing the design of sinogram filters in a straightforward manner. Existing methods include the statistical nonlinear filters [10], bilateral filtering [11], and penalized weighted least squares algorithms [12]. Sinogram data from a commercial scanner, however, are usually unavailable. In addition to that, those methods often suffer from edge blurring or resolution loss.

Iterative reconstruction techniques have attracted a considerable attention over the past decade, especially in the field of LDCT [13]–[15]. Generally speaking, those techniques optimize an objective function that combines the statistical properties of data in the sinogram domain and the prior information in the image domain together. Commonly-used prior information includes non-negativity, maximum CT number, total variation [16], non-local means [17], dictionary learning [18], low rank [19], and their variants. These pieces of generic information can be effectively integrated in the maximum likelihood and compressed sensing frameworks. However, these techniques are time-consuming and require the access to sinogram data and imaging geometry.

Different to the preceding review, image post-processing techniques directly operate on an image that has been reconstructed from raw data and is publicly available (subject to the patient privacy, which can be addressed by anonymizing the images). Traditional image processing methods such as non-local means [17], k-SVD [20], and block-matching 3D [21] improve the image quality to various degrees. However, these techniques may result in uneven performance improvements, potential over-smoothing, and loss of critical subtle structural details.

With the rapid development of deep learning techniques, associated denoising models have achieved an impressive

TABLE I
SUMMARY OF DEEP LEARNING-BASED NETWORK ARCHITECTURE AND THEIR OPTIMIZATION OBJECTIVE FUNCTIONS FOR LDCT DENOISING METHODS. THE ABBREVIATIONS MSE, AL, PL IN TABLE ARE FOR MEAN SQUARED ERROR, ADVERSARIAL LOSS, AND PERCEPTUAL LOSS, RESPECTIVELY.

| Method | Network Architecture | | | Objective Function | | |
|---|---|---|---|---|---|---|
| | Conv | Deconv | Shortcut | MSE | AL | PL |
| **CNN** [5] | ✓ | – | – | ✓ | – | – |
| **RED-CNN** [6] | ✓ | ✓ | Residual Skip | ✓ | – | – |
| **GAN-3D** [7] | ✓ | – | Skip | ✓ | ✓ | – |
| **Cascade-CNN** [8] | ✓ | – | Cascade [1] | ✓ | – | – |
| **WGAN-VGG** [9] | ✓ | – | – | – | ✓ | ✓ |
| **CPCE (Ours)** | ✓ | ✓ | Conveying | – | ✓ | ✓ |

denoising performance for LDCT [5]–[9], [22]. The learning process consists of two key components: network architecture and objective function. The former determines the complexity of the denoising model, while the latter controls how to learn the denoising model. A comprehensive comparison between the proposed method and existing deep learning-based ones is summarized in Table I. The main differences in network architecture and objective function lie in the following two aspects:

**Network architecture**: The denoising model without deconvolutional layers that is the transpose of convolutional layers [23] implies that the input and the output of the denoising model may have different sizes. For example, in the training phase, the input and output sizes of the denoising model in [9] are $80 \times 80$ and $64 \times 64$ respectively. To keep the size of a denoised full-size CT image equal to that of the input, zero-padding in the convolution is needed in the testing phase, which is not used in the training phase. This may lead to inconsistency between training and testing phases and a loss of information. Moreover, the different input and output sizes mildly violate the assumption that $I_{\text{LD}} = I_{\text{ND}} + N$; i.e. the noisy low-dose image $I_{\text{LD}}$ can be expressed as the sum of the reference normal-dose image $I_{\text{ND}}$ and a noise background $N$ [6]. In addition to that, different from the skip connection in [6] that bypasses certain non-linear transformations with an identity function, the conveying path used in the denoising model can reuse early feature-maps as the input to a later layer that has the same feature-map size. This improves the spatial resolution of corresponding features. This direct connection was reported to allow for substantially fewer parameters and a lower computational cost, which, in turn, achieve the state-of-the-art performance in image classification [24].

**Objective function**: Minimizing the MSE based on the difference between the denoised and the normal-dose images led to over-smoothed images [7], [9], which has been shown to correlate poorly with the human perception of image quality [25], [26]. While it is easy to minimize, the optimal MSE estimator suffered from the regression-to-mean problem, which made low-dose denoising results look over-smoothed, unnatural, and implausible. The adversarial loss (AL) in the GAN framework could result in a sharp image locally indistinguishable from the NDCT image but it does not exactly match the NDCT image globally [27] since the AL optimizes the distance between *distributions* of the denoised and NDCT images, rather than a sample-wise distance. The perceptual loss (PL) was hence introduced to make denoised images look more similar to NDCT images in the high-level feature space. Here we combine AL and PL in the same spirit of [9] but in the form of a CPCE formulation to learn from big data more efficiently and in the 3D context to enhance the 2D denoising performance more effectively.

---
[1]This work cascades several CNNs. Different from skip connection and conveying path, these CNNs are trained one by one, not in a unified network.

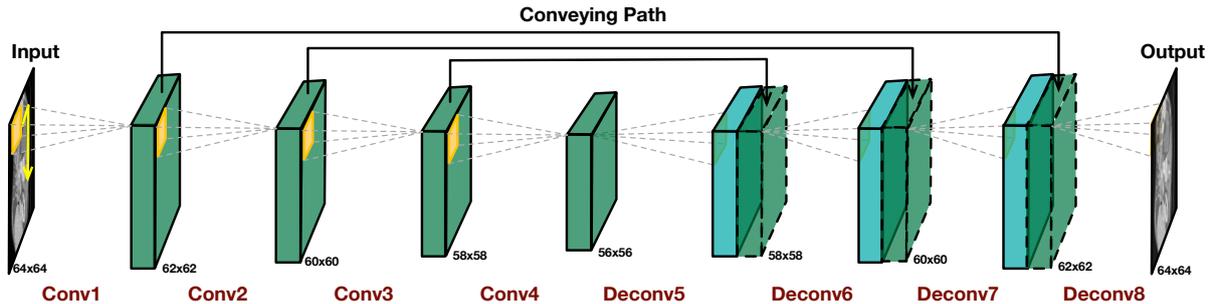

Fig. 1. Our proposed denoising model in the 2D configuration. This network has 4 convolutional and 4 deconvolutional layers. Each layer contains 32 filters, except for the final layer that has only 1 filter. The conveying path copies early feature-maps and reuses them as the input to a later layer with a same feature-map size by the concatenation of feature-maps from the two sides of the conveying-path, which preserves high-resolution features. In order to reduce computational cost, one convolutional layer with $1 \times 1$ filter is used after every conveying path reducing the number of feature-maps from 64 to 32. Each (de)convolutional layer is followed by ReLU. The number next to each feature-map represents its size, given the size of training patches being $64 \times 64$.

### B. Wasserstein GAN framework

Recently, the GAN [25] architecture was developed as a novel way to model the distribution of the given data. A GAN has a pair of neural networks $(G, D)$, where $G$ and $D$ are called the generator and the discriminator, respectively. The generator $G$ takes the input $z$ and generates sample, *i.e.*, $G(z)$. The discriminator $D$ receives samples from both the generator $G$ and the real data $x$, and has to be able to distinguish between the two sources. There is a game relationship between these two networks, where the generator learns to produce more and more realistic samples, and the discriminator learns to become smarter and smarter at distinguishing generated data from real data. These two networks are trained alternatively, and the purpose is that the competition drives the generated samples to be hardly indistinguishable from real data. In the original GAN, the input $z$ is a noise variable sampled from a predefined Gaussian distribution. For the setting of LDCT image denoising task, the input $z$ is taken from LDCT images.

Training the original GAN, however, suffers from several problems such as low quality of generated images, convergence problems, and mode collapse. To address these deficiencies, variants of the GAN were introduced [28], [29], including the Wasserstein GAN (WGAN) [28], which leverages the Wasserstein distance to produce a value function which has better theoretical properties than the original discrepancy measure. A WGAN requires that the discriminator must lie within the space of 1-Lipschitz functions through weight clipping. Replacing the weight clipping with the gradient penalty, the WGAN performs even better [29]. In the LDCT denoising setting, the objective function of WGAN can be described as follows [29]:

$$\min_{\boldsymbol{\theta}_G} \max_{\boldsymbol{\theta}_D} \left\{ \underbrace{\mathbb{E}_{\boldsymbol{I}_{\text{LD}}}\left[D(G(\boldsymbol{I}_{\text{LD}}))\right] - \mathbb{E}_{\boldsymbol{I}_{\text{ND}}}\left[D(\boldsymbol{I}_{\text{ND}})\right]}_{\textbf{Wasserstein distance}} + \lambda \underbrace{\mathbb{E}_{\bar{\boldsymbol{I}}}\left[\left(\|\nabla D(\bar{\boldsymbol{I}})\|_2 - 1\right)^2\right]}_{\textbf{gradient penalty}} \right\}. \quad (1)$$

Here, $\mathbb{E}_a[b]$ denotes the expectation of $b$, as a function of $a$, $\boldsymbol{\theta}_G$ and $\boldsymbol{\theta}_D$ are the parameters of $G$ and $D$, respectively, $\bar{\boldsymbol{I}} = \epsilon \cdot G(\boldsymbol{I}_{\text{LD}}) + (1-\epsilon) \cdot \boldsymbol{I}_{\text{ND}}$ with $\epsilon$ uniformly sampling from an interval of $[0, 1]$. $\nabla D(\bar{\boldsymbol{I}})$ denotes the gradient of $D$ with respect to variable $\bar{\boldsymbol{I}}$. The parameter $\lambda$ controls the trade-off between the Wasserstein distance and the gradient penalty term. The literature has suggested to determine the optimum of (1) by optimizing the generator $G$ and discriminator $D$ iteratively [25], [28], [29], which we employ in this paper.

### III. 2D CPCE NETWORK & 3D EXTENSION

This section presents the proposed 2D CPCE network for LDCT denoising, its extension to a 3D CPCE formulation, and a simple yet effective transfer learning strategy that initializes a 3D model from a 2D trained model. This work includes a unique 2D CNN architecture and a subsequent reconfiguration into a 3D counterpart. Conceptually speaking, a trained 2D network reflects our knowledge on how to denoise an image. Through the transfer learning strategy, this knowledge is efficiently utilized to initialize the corresponding 3D network which, initially, achieves the same denoising performance as the 2D trained model. Finally, we introduce two loss functions to optimize the denoising models.

### A. Proposed denoising model

Assume that $\boldsymbol{I}_{\text{LD}} \in \mathbb{R}^{w \times h}$ is a LDCT image of size $w \times h$, and $\boldsymbol{I}_{\text{ND}} \in \mathbb{R}^{w \times h}$ is the corresponding NDCT image, the relationship can be expressed as follows:

$$\boldsymbol{I}_{\text{LD}} = \mathcal{N}(\boldsymbol{I}_{\text{ND}}), \quad (2)$$

where $\mathcal{N} : \mathbb{R}^{w \times h} \to \mathbb{R}^{w \times h}$ denotes the corrupting process due to the quantum noise that contaminates the NDCT image. The LDCT denoising model/scheme is to provide an approximate inverse $G \approx \mathcal{N}^{-1}$ in order to estimate $\boldsymbol{I}_{\mathrm{ND}}$ from $\boldsymbol{I}_{\mathrm{LD}}$, *i.e.*:

$$G(\boldsymbol{I}_{\mathrm{LD}}) = \boldsymbol{I}_{\mathrm{est}} \approx \boldsymbol{I}_{\mathrm{ND}}. \qquad (3)$$

The proposed network architecture for LDCT is illustrated in Fig. 1, which is referred to as **C**onveying **P**ath-based **C**onvolutional **E**ncoder-decoder (CPCE) network. The CPCE has 4 convolutional layers with all 32 filters, followed by 4 deconvolutional layers also with all 32 filters, except for the final layer that has only 1 filter. A $3 \times 3$ filter with a filter stride 1 is used for all convolutional and deconvolutional layers. The conveying path, originally introduced in the U-net [30] for biomedical image segmentation, copies the early feature-maps and reuses them as the input to a later layer with the same feature-map size, by the concatenation of feature-maps from the two sides of the conveying-path. This, in turn, preserves details of the high-resolution features. Remarkably, dense convolutional networks (DenseNet) [24] receives the feature-maps of all preceding layers in the block fashion, achieving the state-of-the-art classification performance on ImageNet. In this study, the denoising CPCE network has three conveying paths, copying the output of an early convolutional layer and reusing it as the input to a later deconvolutional layer of the same feature-map size. To reduce the computational cost, one convolutional layer with a $1 \times 1$ filter is used after every conveying path reducing the number of feature-maps from 64 to 32. Each convolutional or deconvolutional layer is followed by a rectified linear unit (ReLU). The size of the receptive field in the proposed model is $17 \times 17$.

There are several existing designs that are similar to the CPCE denoising network [6], [30]–[32]. Chen *et al.* proposed a denoising model using a convolutional encoder-decoder network with the skip connection that bypasses the non-linear transformation with an identity mapping [6]. However, the conveying path naturally integrates the properties of identity mappings, deep supervision, and diversified depth, as argued in [24]. With such a path, feature-maps can be reused for later layers, which produces a more effective and compact network that can be trained more efficiently. This leads to a more accurate performance than a less sophisticated network. Other related work includes the U-net [30] and its variants [31], [32]. Compared with these networks, the proposed network does not have pooling or down-sampling layers which could cause information loss. It should be noted that the pooling layer is usually used to reduce the spatial dimension and gain (small) translation-invariance [33], especially in the representation learning for image classification.

### B. 3D Spatial information from adjacent LDCT slices

Most of the existing denoising networks focus on image denoising in 2D. However, the adjacent image slices in a CT volume have strong correlative features that can potentially improve the performance of image denoising. Such a spatial synergy is routinely used in radiologists' image reading when they step through a stack of image slices or view these slices via volumetric rendering. Therefore, we propose to incorporate 3D spatial information from adjacent slices for LDCT denoising.

Since the spatial correlation between the input slice and its adjacent slices is strong, we expand a single 2D slice input of the 2D denoising network to include its adjacent slices. Here we take its adjacent two slices as an example; *i.e.*, including upper and lower LDCT image slices. With the expanded input of three LDCT slices together, the original 2D convolutional filter should be extended to a 3D convolutional filter. Fig. 2 presents the change in the first convolutional layer of the proposed denoising network.

- **Input**: Augment one LDCT slice with two adjacent LDCT slices as an extended dimension, *i.e.*, depth;
- **Filter**: Replace a 2D convolution of size $3 \times 3$ with a 3D convolution of size $3 \times 3 \times 3$.

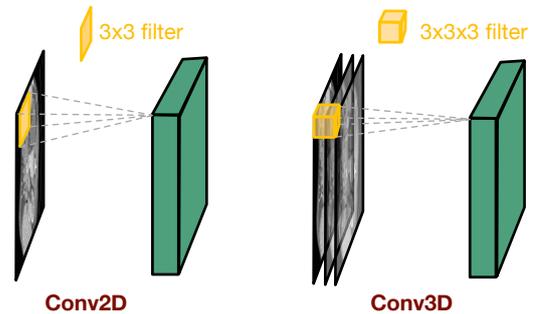

Fig. 2. Transfer learning from a 2D filter to a 3D counterpart. Left plot is the 2D convolutional filter of size $3 \times 3$ with single one input LDCT image, right one is the 3D convolutional filter of size $3 \times 3 \times 3$ with three input LDCT images. Both resultant feature-maps are in the same size since the depth dimension of feature-maps on right is reduced to 1.

With more than 3 input slices, this 3D extension is applied to later convolutional layers of the 2D denoising model until the dimension of the depth is reduced to 1. However, with more input slices, the two sides of a conveying-path may have different feature-map sizes; *i.e.*, the feature-map on the left-hand side is a 4D-tensor in terms of [*width, height, depth, channel*] while the feature-map on the right-hand side is a 3D-tensor in terms of [*width, height, channel*], which implies that the two sides cannot be concatenated. To address this issue, the conveying path only copies features at the middle location of the depth on the left-hand side that are most related to the output image and reused them on the right-hand side to preserve structural details.

We refer to our 2D denoising network as CPCE-2D and the resultant hybrid 2D/3D denoising network as CPCE-3D.

### C. Transfer learning strategy

One brute-force method is to train CPCE-3D denoising model from scratch. Generally, such a direct training process is computationally expensive. Conversely, the CPCE-2D model has less parameters and is, therefore, easier to train compared to the CPCE-3D model. With the availability of a trained CPCE-2D model, the following interesting problem emerges: *how to utilize the 2D network to initialize the CPCE-3D model instead of random initialization?* Our answer to this lies in examining the following strategy.

This strategy is a transfer learning idea that extends the parameter setting of the 2D filter to the corresponding 3D

filter along with complementary zero padding. More formally, assume $\boldsymbol{H} \in \mathbb{R}^{c_{\text{in}} \times c_{\text{out}} \times 3 \times 3}$ is a trained 2D convolutional filter where $c_{\text{in}}$ and $c_{\text{out}}$ denote the channel numbers of the input and output to this filter, respectively, then the corresponding 3D convolutional filter $\boldsymbol{B} \in \mathbb{R}^{c_{\text{in}} \times c_{\text{out}} \times 3 \times 3 \times 3}$ can be initialized as follows:

$$\begin{cases} \boldsymbol{B}_{(0)} & = \mathbf{0}_{c_{\text{in}} \times c_{\text{out}} \times 3 \times 3} \\ \boldsymbol{B}_{(1)} & = \boldsymbol{H}_{c_{\text{in}} \times c_{\text{out}} \times 3 \times 3} \\ \boldsymbol{B}_{(2)} & = \mathbf{0}_{c_{\text{in}} \times c_{\text{out}} \times 3 \times 3}, \end{cases} \quad (4)$$

where the subscript of $\boldsymbol{B}$ corresponds to the depth of 3D convolution, *i.e.*, the last dimension of $\boldsymbol{B}$. By design, the initialized 3D filter is identical to a 2D filter, and consequently the initial 3D denoising model, based on a trained 2D model, can achieve the same performance as the trained 2D model. Then, the fine tuning process helps to utilize the 3D spatial context and enhance the 2D denoising performance. Compared to optimizing the 3D network after a random initialization, the transfer learning strategy can leverage the trained 2D model for accelerated training and an improved chance of producing a better denoising performance.

Note that our 3D network is different from the 3D network in [7] because 1) our network is a hybrid 2D/3D network which has a 2D output, while the model in [7] is a pure 3D network whose output is also a 3D volume. Moreover, our hybrid 2D/3D network is a trade-off between a 2D network and a 3D network. This implies that our network architecture has not only fewer parameters than a pure 3D network but also preserves the 3D spatial information that is ignored by a 2D network; and 2) the proposed transfer learning strategy can achieve a faster convergence and train a better model than training from scratch, which is also beneficial to the 3D network in [7].

### D. Objective function

Inspired by the impressive results in [7], [9], this paper optimizes the denoising network in the WGAN framework with two loss functions as follow.

**Adversarial loss**: Adversarial loss encourages the generator network to produce samples that are indistinguishable from the NDCT images, which refers to the loss function of the generator in (1) [29]:

$$\min_{\boldsymbol{\theta}_G} \mathcal{L}_a = \mathbb{E}_{\boldsymbol{I}_{\text{LD}}} \left[ D\big(G(\boldsymbol{I}_{\text{LD}})\big) \right], \quad (5)$$

where the last two terms in (1) are constant with respect to $\boldsymbol{\theta}_G$. The proposed denoising network is the generator $G$ in the GAN framework. The discriminator $D$ used in this paper has 6 convolutional layers with 64, 64, 128, 128, 256, and 256 filters, followed by 2 fully-connected layers of sizes 1024 and 1. Each layer is followed by a leaky ReLU, which has a negative slope of 0.2 when the unit is saturated and not active. A $3 \times 3$ filter is used for all convolutional layers. A unit filter stride is used for odd convolutional layers and this stride is doubled for even layers.

**Perceptual loss**: The perceptual similarity measure, proposed in [34], [35], computes the distance between $G(\boldsymbol{I}_{\text{LD}})$ and $\boldsymbol{I}_{\text{ND}}$ in a high-level feature space by a differential function $\phi$, rather than in the pixel space. This assessment allows the denoising model to produce denoised images that may not match the NDCT image with pixel-wise accuracy, but drives the network to generate images that have a visually desirable feature representation to aid radiologists optimally. Following the ideas described in [9], [27], [34], [35], here we choose a pre-trained VGG-19 network [36] as the feature map $\phi$. In the experiments, the feature map $\phi$ takes the 16th convolutional layer in the VGG network. The perceptual loss is then defined as:

$$\min_{\boldsymbol{\theta}_G} \mathcal{L}_p = \mathbb{E}_{(\boldsymbol{I}_{\text{LD}}, \boldsymbol{I}_{\text{ND}})} \| \phi\big(G(\boldsymbol{I}_{\text{LD}})\big) - \phi\big(\boldsymbol{I}_{\text{ND}}\big) \|_2^2. \quad (6)$$

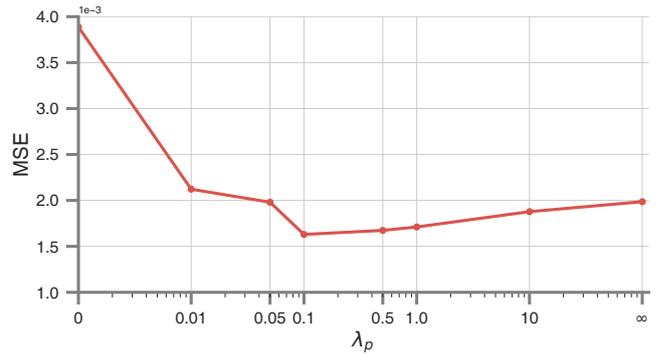

Fig. 3. The effect of the parameter $\lambda_p$ in the CPCE-2D model on the validation set from the Mayo dataset. Note that parameter $\lambda_p = 0$ ($\lambda_p = \infty$) indicated that the denoising model was only optimized with respect to the adversarial loss (perceptual loss).

The final objective function for optimizing the proposed denoising network is defined as follows:

$$\min_{\boldsymbol{\theta}_G} \mathcal{L} = \mathcal{L}_a + \lambda_p \mathcal{L}_p. \quad (7)$$

Since our final objective function has an additional perceptual loss term, the similarity is encouraged between generated images and the NDCT images in the high-level feature space. In addition to that, the inclusion of the adversarial loss enhances textural information in the denoised images.

## IV. EXPERIMENTS

This section presents the experimental setup and describes the denoising performance of the introduced CPCE method with recently published competitive methods on simulated and real low-dose CT datasets. Note that we use CPCE-3D($i$) to denote that the number of input slices of the CPCE-3D model is $i$, and the proposed training learning scheme by a superscript "+" mark (*e.g.*, CPCE-3D(9)$^+$).

### A. Low-dose dataset with simulated noise

The experimental data set stems from an authorized clinical low-dose CT dataset, which was made for *the 2016 NIH-AAPM-Mayo Clinic Low-Dose CT Grand Challenge*[2]. This dataset included normal-dose abdominal CT images that were taken from 10 anonymous patients and the corresponding

---
[2]http://www.aapm.org/GrandChallenge/LowDoseCT/

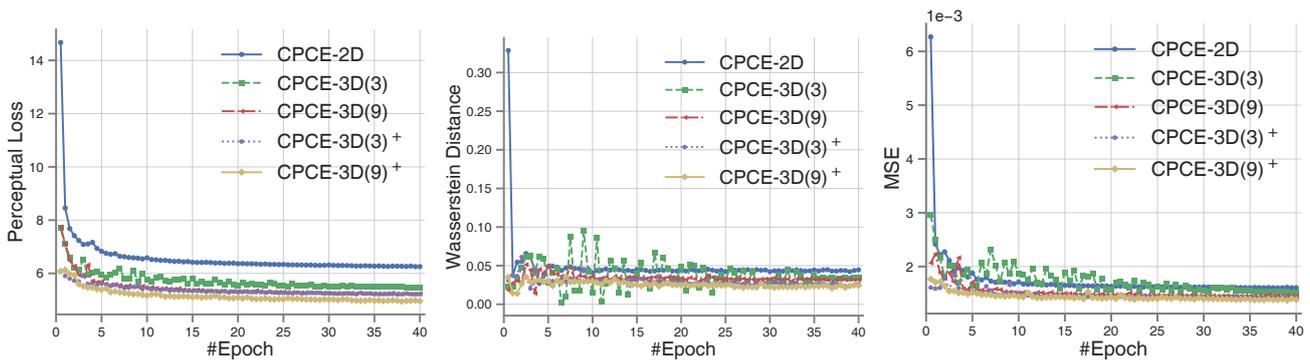

Fig. 4. Comparisons between training with transfer learning and training from scratch in terms of the perceptual loss, Wasserstein distance, and MSE on the Mayo dataset. The 3D model trained with transfer learning was marked by a superscript $+$. Note that the smaller the better for all these three metrics.

*simulated* quarter-dose CT images. The slice thickness and reconstruction interval in this dataset were 1.0 mm and 0.8 mm, respectively.

*1) Experimental setting:* For training purposes, 128K image patches of size $64 \times 64$ were *randomly* selected from 5 patients that were *randomly* selected from this dataset. To validate the performance of the trained models, 64K image patches were *randomly* selected from the remaining 5 patients. For the CPCE-3D denoising method, the adjacent low-dose image patches were kept for training and testing. In addition to that, the dynamic range of the CT image was first normalized into the unit interval $[0, 1]$ for the training of the neural network, and then rescaled to the interval $[0, 255]$ for the use of the VGG network. Since the VGG network was trained with natural images, each CT image was copied three times to serve as the RGB channels of the VGG network.

*2) Parameter setting:* During the training phase, the Adam optimization method [37] was used to train the CPCE denoising model with a mini-batch of 128 image patches for each iteration. For the training from scratch, the learning rate $\alpha$ was selected to be $1.0 \times 10^{-4}$ with two exponential decay rates $\beta_1 = 0.9$ and $\beta_2 = 0.999$ for the moment estimates. However, the learning rate halved for training based on transfer learning, which was followed by fine-tuning. In all the experiments, the learning rate was adjusted by $1/t$ decay; namely, $\alpha_t = \alpha/t$ at the $t$-th epoch. The trade-off parameter $\lambda$ between the Wasserstein distance and the gradient penalty was set to be 10, which was suggested in [29]. All networks were implemented in the TensorFlow Library [38] and trained with a NVIDIA Titan Xp GPU.

In order to determine the weighting parameter $\lambda_p$ for the perceptual loss in the objective function, we selected the parameter $\lambda_p$ from $\{0, 0.01, 0.05, 0.1, 0.5, 1.0, 10, \infty\}$. Note that parameter $\lambda_p = \infty$ ($\lambda_p = 0$) indicated that the denoising model was only optimized with respect to the perceptual loss (adversarial loss); otherwise, the denoising model was optimized by balancing these two losses. When varying $\lambda_p$, it was impossible to use the perceptual loss and Wasserstein distance as metrics, because 1) a larger $\lambda_p$ was associated with a smaller perceptual loss on the validation set; and 2) the Wasserstein distance was not informative when the denoising model was only optimized with the perceptual loss. Therefore, we used the mean squared error (MSE) as the metric to measure the denoising performance on the validation set, as shown in Fig. 3. Note that we chose the CPCE-2D model as the baseline here to select $\lambda_p$ since it has fewer parameters and is more efficient to train multiple times.

The results demonstrated that 1) parameter $\lambda_p = 0.1$ achieved the lowest MSE, which was used in the following experiments; and 2) compared to the adversarial loss, the perceptual loss had a dominant influence on the denoising performance.

*3) Convergence behavior:* Here, we discussed the convergence differences between training from scratch and training through the proposed transfer learning strategy when given 3 and 9 input slices for the CPCE-3D models. This involved evaluating both denoising models at every half epoch on the validation set with the perceptual loss, Wasserstein distance, and MSE as shown in Fig. 4. The perceptual loss computed the content similarity between the denoised and NDCT patches in a high-level feature space while MSE did this in the pixel space. The Wasserstein distance corresponded to the performance quality of the GAN framework [28], [29]. Note that absolute Wasserstein distance was provided in the testing phase since it can be either negative or positive in the objective function (1). In this experiment, we used a trained CPCE-2D model at epoch 10 for transfer learning where the downward trend of the validation error was not strong; also see our supplementary for results with 5 and 7 input slices.

Fig. 4 highlights that the CPCE-3D model, based on a trained CPCE-2D model, achieved lower perceptual loss, MSE, and Wasserstein distance than the counterpart model trained from scratch, showing that the 3D denoising model can be trained more accurately with the proposed transfer learning strategy. With initial weights from the 2D trained model, the fine-tuning technique utilized the 3D spatial information to enhance the denoising performance.

In general, the 2D trained model was assumed to be available and did not need to be retrained from scratch. Also, the time saved from transfer learning depends on how many adjacent slices are included. In this study, the CPCE-3D denoising model required 17 minutes and 32 minutes to compute half the number of iterations per epoch for 3 and 9 input slices respectively, whilst the CPCE-2D denoising model only required 14 minutes. In the case of the number of input slices being 9, the same 3D network trained through transfer learning only required approximately 10 epochs to achieve the same denoising performance in term of perceptual loss that the 3D network trained from scratch achieved with 40 epochs, which implies our transfer learning strategy saved 75% computational time in this case. If we include the computational time taken by the trained 2D model, the 3D network trained through transfer learning required 10 epochs for 2D training and 10 epochs for 3D training to achieve the same denoising performance, which means that our transfer learning strategy saved about 65% computational time. The savings could be more significant when the number of slices included is even larger or our transfer learning strategy could be further refined. More importantly, our experimental results show that the training process, which utilizes transfer learning, converged faster and achieved a better denoising performance than the networks trained from scratch. In brief, our results suggest that the transfer learning approach is not only more efficient but also more effective in dealing with a non-convex optimization problem.

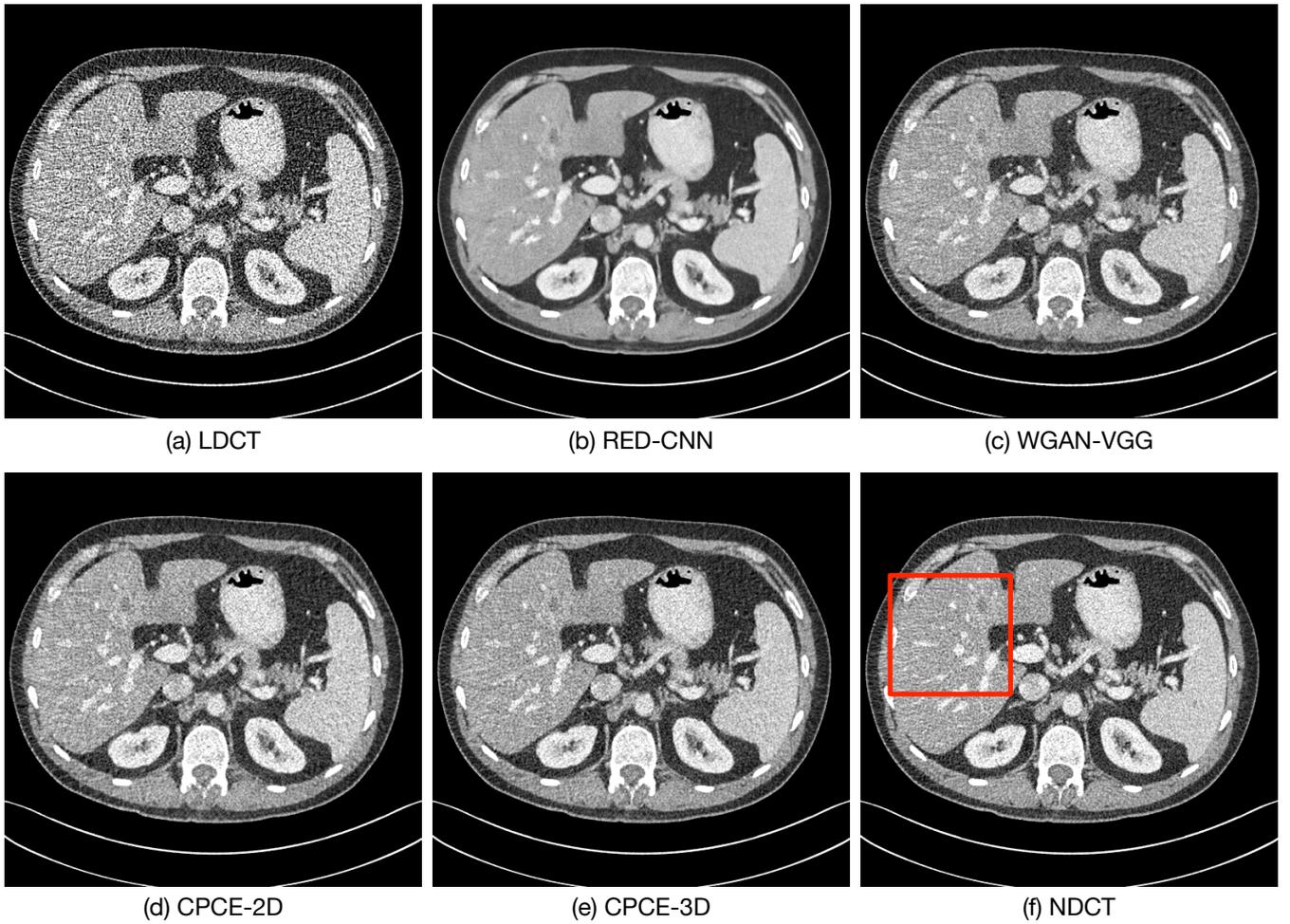

Fig. 5. Transverse CT image through the abdomen. ROI is shown in the red rectangle. The display window of this slice is $[-180, 200]$ HU for better visualization of the lesion.

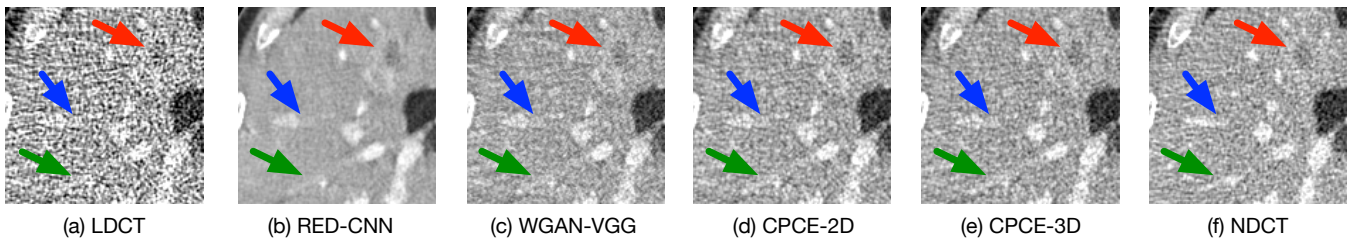

Fig. 6. Zoomed ROI of the red rectangle in Fig. 5. The red arrow indicates low attenuation lesion in the left lobe of liver, the blue and green arrows indicate intrahepatic blood vessels. The display window is $[-180, 200]$ HU.

*4) Denoising performance:* To visualize the denoising performance, we selected two exemplary slices that contain low attenuation lesions and blood vessels for the clinical task-based assessment, as shown in Figs. 5 and 7. The purpose of these figures was to compare our proposed CPCE-2D and CPCE-3D denoising networks with some of the latest published networks, including RED-CNN [6] and WGAN-VGG [9]. We used CPCE-3D(9)$^+$ as the best CPCE-3D model for visualization since it achieved the best denoising performance among the CPCE-3D models shown in Fig. 4. Compared with the NDCT image, RED-CNN produced over-smoothed results that led to a loss of texture information. This was a direct result of the MSE-based optimization suffering from the regression-to-mean problem.

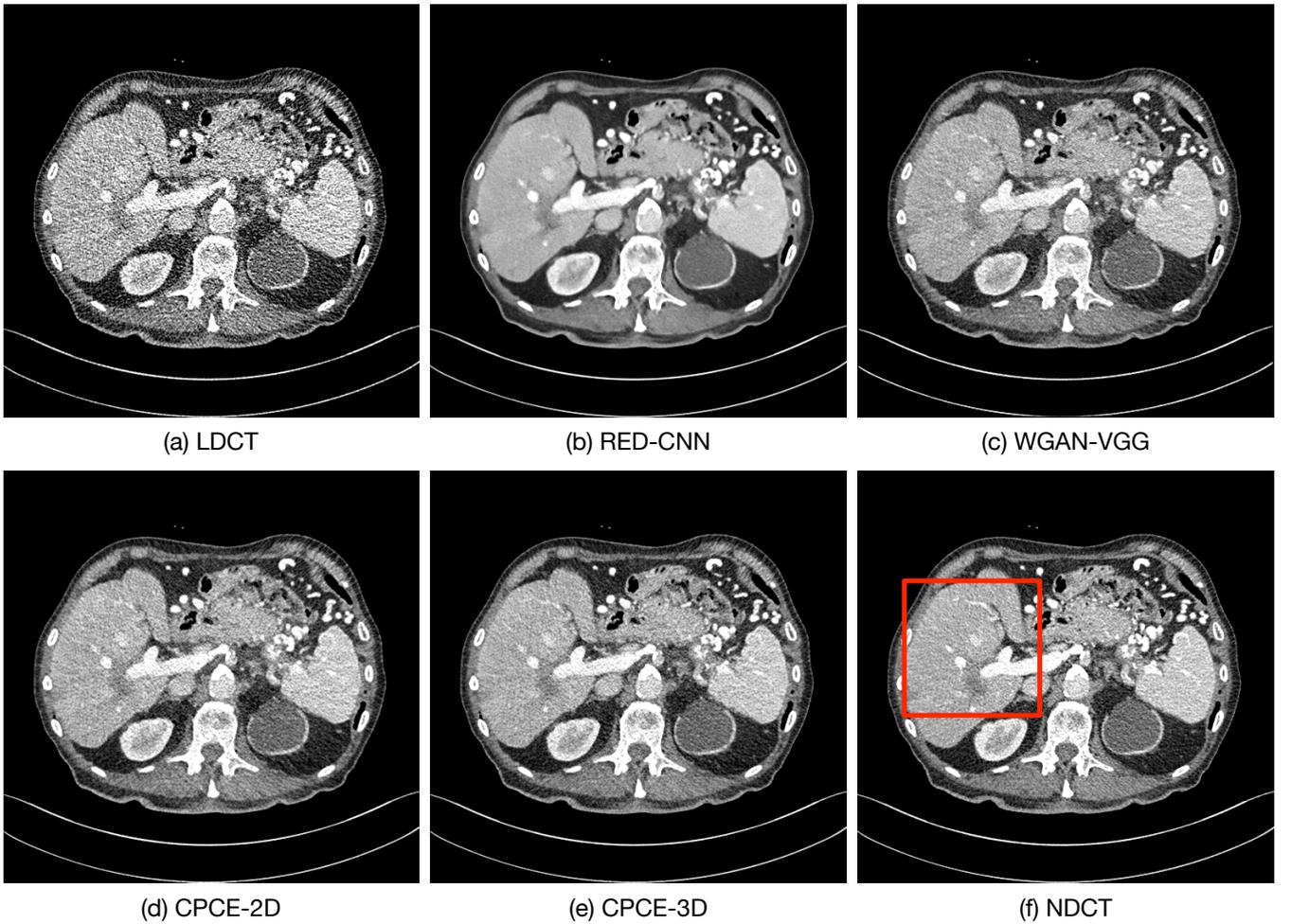

Fig. 7. Transverse CT image through the abdomen. ROI is shown in the red rectangle. The display window is $[-160, 240]$ HU for better visualization of the lesion.

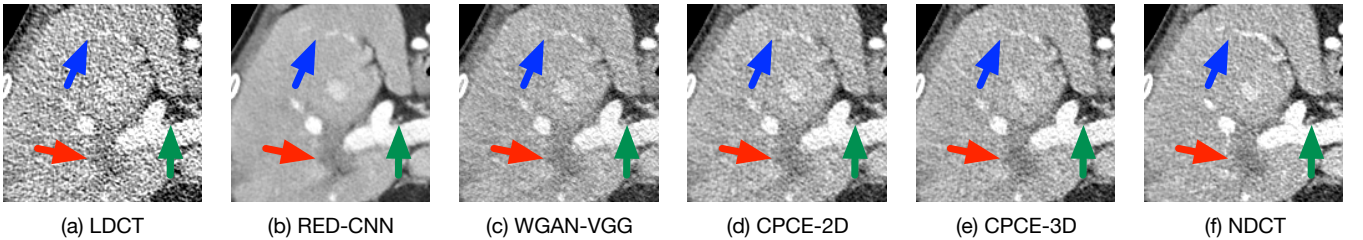

Fig. 8. Zoomed ROI of the red rectangle in Fig. 7. The red arrow indicates the low attenuation lesion in the posterior right liver lobe, the blue and green arrows indicate the blood vessels of the liver. The display window is $[-160, 240]$ HU.

Our proposed CPCE-2D and CPCE-3D denoising net-

TABLE II
RELATIONSHIP AMONG FOUR METRICS EVALUATING IMAGE QUALITY.

| Metric \ Domain Similarity | Pixel space | Feature space |
|---|---|---|
| Content | **PSNR** | **PL** |
| Texture | **SSIM** | **TML** |

works produced denoising results comparable to that from the WGAN-VGG network because they were optimized in the same framework. When ROIs were focused on, however, the CPCE networks improved the denoising performance relative to the WGAN-VGG and RED-CNN models in Figs. 6 and 8. The low attenuation lesion (indicated by the red arrow) was clearly visualized with all networks as it is clear in the low-dose CT images, but the blood vessels (by the green and blue arrows) were best preserved with the CPCE-3D network, as confirmed by the NDCT benchmark. In other words, our CPCE-3D did achieve a better denoising result for visualizing the low attenuation lesion and revealing the blood vessels compared to other methods. However, it is important to note that the competing techniques could also identify the blood vessel but not up to the same clarity as our method can.

TABLE III
COMPARISON AMONG DIFFERENT METHODS ON THE MAYO DATASET. THESE FOUR METRICS ARE PEAK SIGNAL-TO-NOISE RATIO (PSNR), STRUCTURE SIMILARITY (SSIM), PERCEPTUAL LOSS (PL), AND TEXTURE MATCHING LOSS (TML). 3D NETWORK THAT TRAINED THROUGH TRANSFER LEARNING IS MARKED WITH $^+$. THE RESULTS ARE SHOWN IN FORM OF $mean(std)$ BASED ON ALL SLICES IN THE TESTING SET. FOR EACH METRIC, WE MARK THE BEST IN RED THE SECOND BEST IN BLUE.

| Metric | LDCT | RED-CNN | WGAN-VGG | CPCE-2D | CPCE-3D(3) | CPCE-3D(5) | CPCE-3D(7) | CPCE-3D(9) | CPCE-3D(3)$^+$ | CPCE-3D(5)$^+$ | CPCE-3D(7)$^+$ | CPCE-3D(9)$^+$ |
|---|---|---|---|---|---|---|---|---|---|---|---|---|
| PSNR | 26.073 (2.219) | **31.390** (**1.849**) | 28.876 (1.620) | 29.620 (1.857) | 29.838 (1.846) | 29.995 (1.897) | 30.008 (1.878) | 30.045 (1.888) | 30.002 (1.883) | 30.037 (1.929) | 30.136 (1.904) | **30.137** (**1.938**) |
| SSIM | 0.834 (0.059) | **0.919** (**0.032**) | 0.896 (0.039) | 0.898 (0.039) | 0.900 (0.038) | 0.902 (0.038) | 0.903 (0.038) | 0.903 (0.038) | 0.903 (0.038) | 0.903 (0.038) | **0.905** (**0.038**) | **0.905** (**0.038**) |
| PL | 4.81 (1.18) | 4.31 (0.90) | 2.55 (0.74) | 2.37 (0.58) | 2.06 (0.51) | 1.99 (0.50) | 1.96 (0.50) | 1.95 (0.49) | 2.02 (0.51) | 1.90 (0.49) | **1.88** (**0.48**) | **1.85** (**0.47**) |
| TML | 258.69 (133.42) | 180.51 (87.99) | 83.07 (39.04) | 82.96 (36.53) | 73.68 (34.01) | 71.61 (33.07) | 70.89 (32.50) | 70.79 (32.58) | 72.11 (33.71) | 69.10 (32.61) | **68.46** (**32.37**) | **67.63** (**32.05**) |

Note that the normal-dose and the corresponding low-dose CT images of Mayo data are in perfect registration, and were used to train our network. The objective of both CPCE-2D and CPCE-3D networks was not only to reduce noise in the LDCT images, but also to preserve the textural information in the NDCT images. This, in turn, allows providing radiologists with a better image background for multiple clinical diagnosis tasks, not limited to the lesion detection task, even though the lesion detection task is the main purpose of the Mayo dataset.

For a quantitative comparison of denoised full-size CT slices, we used all slices in the testing set to compare the denoising performance. In addition to the peak signal-to-noise ratio (PSNR), structural similarity (SSIM) [26], and PL used in the objective function, we also employ the texture matching loss (TML) to measure the texture similarity in a high-level feature space [27]. Gatys *et al.* demonstrated how convolutional neural networks can be used to create a high-quality texture [39], [40]. Given a target texture image, the output image can be iteratively generated by matching statistics extracted from a pre-trained network to the target texture. The mapping $\phi(\boldsymbol{I}) \in \mathbb{R}^{n \times m}$ at a given VGG layer that has $n$ features of length $m$ is used to compute the texture matching loss:

$$\mathcal{L}_t = \|GM(\phi(\boldsymbol{I}_{\text{est}})) - GM(\phi(\boldsymbol{I}_{\text{ND}}))\|_2^2 \quad (8)$$

based on the Gram matrix $GM(\boldsymbol{F}) = \boldsymbol{F}\boldsymbol{F}^{\text{T}} \in \mathbb{R}^{n \times n}$. In order to measure the local texture similarity between the denoised and normal-dose images, we computed the texture matching loss $\mathcal{L}_t$ patch-wise as suggested in [27], where the full-size CT images were divided into patches of size $64 \times 64$. The feature map $\phi$ was the same as what we used in the perceptual loss. Table II summarizes the relationship among these four metrics used in this study.

Table III presents a quantitative comparison of the denoising performance, showing that

1) RED-CNN achieved the highest PSNR as a result of the MSE-based objective function. However, the highest PSNR values did not guarantee that the denoised images have a best perceptual and texture similarity with the NDCT images in the high-level feature space as measured by PL and TML. As Figs. 5 and 7 show, the RED-CNN produced over-smoothed images and a loss of texture information, resulting from the regression-to-mean problem.
2) The proposed CPCE-2D performed better than the WGAN-VGG as confirmed by these four metrics, indicating the advantage of the proposed network architecture.
3) Comparing the CPCE-3D model with the 2D counterpart, the 3D spatial information from adjacent slices helped the CPCE-3D model improve the denoising performance.
4) For CPCE-3D models, the transfer learning strategy consistently achieved a better denoising performance when compared to a training from scratch for different numbers of input slices. Moreover, increasing in the number of input slices produces an improvement in the denoising performance of the CPCE-3D network.

Overall, the proposed CPCE-3D model, based on a trained 2D model, achieved the best performance in suppressing image noise and preserving subtle structures among all methods used in this comparison, as shown in Fig. 5, Fig. 7, and Table III.

### B. Low-dose dataset with real noise

We also validated our method on a real low-dose CT dataset, the Massachusetts General Hospital (MGH) dataset [41], which contains 40 cadaver scans acquired with representative protocols. Each cadaver was scanned with a GE Discovery HD750 scanner at 4 different dose levels; *i.e.*, $10NI$, $20NI$, $30NI$, and $40NI$. Here $NI$ (Noise Index) is referenced to the standard deviation of CT numbers within a region of interest in a water phantom of a specific size [42], which was used by GE as a measurement of the dose level. In our experiments, we used FBP images reconstructed from the $40NI$ dataset as the low-dose input and the FBP images reconstructed from $10NI$ as the normal-dose label. It is important to note that the low-dose CT images and corresponding normal-dose CT images were not in perfect registration due to the error in the patient table re-positioning and the uncertainty in the source angle initialization.

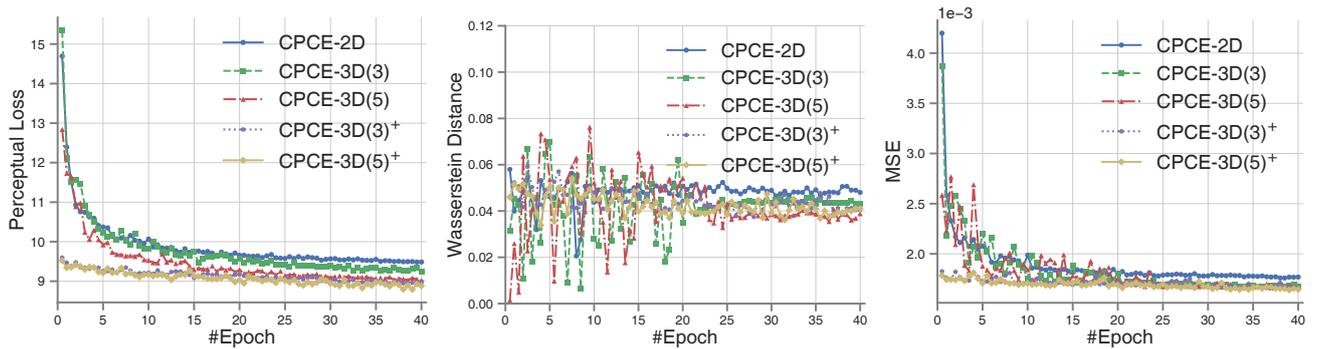

Fig. 9. Comparisons between training with transfer learning and training from scratch in terms of perceptual loss, Wasserstein distance, and MSE on the MGH dataset. The 3D model trained with transfer learning was marked by a superscript $+$. Note that the smaller the better for all these three metrics.

TABLE IV
COMPARISON AMONG DIFFERENT METHODS ON MGH DATASET. THESE FOUR METRICS ARE PEAK SIGNAL-TO-NOISE RATIO (PSNR), STRUCTURE SIMILARITY (SSIM), PERCEPTUAL LOSS (PL), AND TEXTURE MATCHING LOSS (TML). 3D NETWORK THAT TRAINED THROUGH TRANSFER LEARNING IS MARKED WITH $+$. THE RESULTS ARE SHOWN IN FORM OF $mean \pm std$ BASED ON ALL SLICES IN THE TESTING SET. FOR EACH METRIC, WE MARK THE BEST IN RED THE SECOND BEST IN BLUE.

| Metric | LDCT | RED-CNN | WGAN-VGG | CPCE-2D | CPCE-3D(3) | CPCE-3D(5) | CPCE-3D(3)$^+$ | CPCE-3D(5)$^+$ |
|---|---|---|---|---|---|---|---|---|
| PSNR | 25.354±1.946 | **31.424±1.826** | 28.393±1.553 | 29.362±1.603 | 29.588±1.684 | 29.590±1.668 | 29.606±1.682 | 29.631±1.686 |
| SSIM | 0.823±0.060 | **0.937±0.032** | 0.905±0.038 | 0.905±0.038 | 0.906±0.039 | 0.906±0.038 | 0.906±0.038 | 0.906±0.038 |
| PL | 7.29±1.70 | 6.22±1.46 | 3.74±0.75 | 3.67±0.74 | 3.55±0.76 | 3.48±0.76 | 3.46±0.73 | **3.44±0.72** |
| TML | 383.18±160.10 | 182.60±68.22 | 113.61±40.54 | 111.76±38.97 | 109.98±38.59 | 107.52±37.06 | 106.08±36.76 | **105.99±36.75** |

TABLE V
SUBJECTIVE QUALITY SCORES ($mean \pm std$) FOR DIFFERENT METHODS INVOLVED.

| | NDCT | LDCT | RED-CNN | WGAN-VGG | CPCE-2D | CPCE-3D |
|---|---|---|---|---|---|---|
| | | | **Mayo dataset** | | | |
| Noise Removal | - | - | **4.70 ± 0.16** | 3.00 ± 0.22 | 3.15 ± 0.21 | 3.45 ± 0.18 |
| Artifact Reduction | - | - | 3.05 ± 0.17 | 3.35 ± 0.21 | 3.50 ± 0.21 | **3.60 ± 0.13** |
| Structure Preservation | - | - | 2.80 ± 0.28 | 2.95 ± 0.20 | 3.05 ± 0.20 | **3.45 ± 0.12** |
| Overall quality | 4.00 ± 0.15 | 1.25 ± 0.15 | 3.25 ± 0.17 | 3.55 ± 0.20 | 3.65 ± 0.20 | **3.85 ± 0.17** |
| | | | **MGH dataset** | | | |
| Noise Removal | - | - | **4.50 ± 0.19** | 3.45 ± 0.17 | 3.55 ± 0.17 | 3.75 ± 0.18 |
| Artifact Reduction | - | - | 2.90 ± 0.31 | 3.05 ± 0.16 | 3.20 ± 0.16 | **3.30 ± 0.16** |
| Structure Preservation | - | - | 2.10 ± 0.34 | 3.10 ± 0.19 | 3.20 ± 0.18 | **3.35 ± 0.07** |
| Overall quality | 4.05 ± 0.20 | 1.40 ± 0.17 | 2.40 ± 0.33 | 3.55 ± 0.20 | 3.60 ± 0.20 | **3.70 ± 0.13** |

We randomly pre-selected the CT scans of 8 cadavers, whose cross-plane image resolutions were within [0.7031, 0.7343] mm since we did not want the resolution difference to be a potential factor that negatively affects the network training. The slice thickness and reconstruction interval were both 2.5 mm, which were consistent among the whole dataset.

In this experiment, we randomly selected 4 out of the 8 pre-selected cadavers as the training set and the remaining ones for validation purposes. The experimental setup was the same as that applied for denoising the Mayo dataset in the previous section. The initial 3D denoising model was transferred from the CPCE-2D model at epoch 20 where downward trend at this point was not strong. Fig. 9 presents the comparison between training based on transfer learning and training from scratch on the validation set. We only used CPCE-3D with 3 and 5 input slices since the slice thickness was very large. The experimental results concerning this practical application have confirmed similar gains, *i.e.* the network converged faster and achieved better denoising performance through training based on transfer learning compared to a training from scratch. This is in line with the denoising of the simulated Mayo dataset. The curves were not as smooth as that from the Mayo dataset since the low-dose images and normal-dose images were not perfectly registered in the practical scans.

Table IV presents the comparison among all methods on each of the full-size CT images in the testing set. Based on the four metrics, the results confirm the advantages of the proposed CPCE method and the transfer learning strategy observed in the previous subsection. The differences caused by the large slice thickness are 1) the improvement in the pixel space from 2D to 3D was not as significant as what was observed when applied to the simulated Mayo dataset; and 2) the preservation of small structures were not as evident as on this practical dataset; see the supplemental material for the denoising examples which show that the RED-CNN produced aggressively over-smoothened images in this real-dataset.

### C. Blind reader study for Mayo and MGH datasets

For each dataset, we performed a blind reader study on 10 groups of the image slices, which were randomly selected from test patients. Each group contains the LDCT image, the NDCT image, and the denoised LDCT images. In each group, the information about the LDCT and NDCT images was provided as the references for the radiologists to rate the images denoised using different methods. The radiologists were not given any information on which method was utilized for the denoising. Two radiologists were asked to score each denoised image independently in terms of noise removal, artifact reduction, structure preservation, and overall quality on a five-point scale (1 = unacceptable and 5 = excellent). The NDCT image served as the gold standard, and LDCT image as the bottom-line. Since the radiologists did have the same reference (LDCT and NDCT) images, their evaluation scores of the denoised images were calibrated. In order to examine whether they used the five-point scale in an equivalent way, they were also asked to grade the overall quality of the LDCT and NDCT images. The scores from the radiologists were then reported as $mean \pm std$ (averaged score of the two radiologists $\pm$ standard deviations) to obtain the final evaluation results in Table V.

This table shows that the MSE-based objective function gives the best noise removal while methods optimized under the WGAN framework with an additional perceptual loss achieve better scores in terms of the artifact reduction, the structural preservation, and the overall quality. The standard deviations of the reference images were small, which implies that two radiologists evaluated the images very similarly. The CPCE-2D model has a slightly better score than WGAN-VGG model, which suggests an advantage of conveying-path based encode-decode network structure. The CPCE-3D model also has a slightly better score than CPCE-2D, which implies an advantage of incorporating 3D spatial information. In summary, MSE-based network is good at noise removal but at a cost of losing image details, resulting in an image quality degradation for diagnosis. More precisely, the combination of WGAN and 3D structure offers a better overall image quality than the other methods studied here.

## V. Discussion and Conclusion

This article has introduced a novel 2D low-dose CT denoising approach, referred to as the conveying path-based convolutional encoder-decoder (CPCE) network. This approach is featured by 1) a convolutional encoder-decoder network with three conveying-paths, allowing to reuse feature-maps of early layers as the input to later layers, facilitating preservation of high-resolution features, and 2) the Wasserstein GAN framework optimized with an additional perceptual loss. In addition to the conceptually attractive and practically effective CPCE network, a simple yet advantageous strategy has been applied to extend 2D networks to train 3D networks; that is, the 3D network is initialized by a trained 2D model, which can, consequently, achieve a better denoising performance than its 3D counterpart that is directly trained from scratch. It is important to note that the transfer learning from 2D to 3D architectures allows a more stable and more efficient training when compared to the standard generic 3D training. This article has also compared these developed approaches to the recently published RED-CNN and WGAN-VGG networks and has confirmed that the CPCE 2D and 3D networks perform favorably for suppressing image noise without compromising image texture.

On the basis of the results presented here, it is of great interest to improve or generalize the transfer learning scheme. By design, the initialized 3D filter we currently use is identical to the trained 2D filter, and can take full advantage of the 2D training experience. For example, assuming that we have a trained 2D convolutional neural networks in an architecture similar to the 3D network in [7] (*i.e.*, we use 2D convolutions instead of 3D ones in [7]), the 3D network in [7] can be easily initialized based on the trained 2D model and subsequently refined. Importantly, the strategy introduced here is expected to be applicable to all existing 2D models, such as those discussed in [5], [6], [8], [9]. While we have focused on this straightforward transfer learning scheme, it is acknowledged that other 2D to 3D or higher dimensional transfer learning ideas could potentially produce better results.

The work presented in this article results in a number of possibilities for future research. As an example, we can start with three 2D networks trained for transverse, sagittal, and coronal sections, respectively, and then initialize the 3D network. More precisely, the trained 2D models carry information for image recovery and should be useful in guiding or regularizing the 3D model in its training process. Moreover, in addition to utilizing the 3D spatial information from adjacent slices to enhance the 2D performance, it would be also helpful to combine complementary methods/networks through bagging/ensemble learning for a best denoising performance [43], [44]. As an example, given all trained models, the combination of different outputs may also further improve the denoising performance. It should also be noted that Zia *et al.* proposed to extend a 2D filter to a 3D one by replicating this 2D filter along the depth dimension for RGB-D object recognition [45]. However, that extension cannot guarantee that the initial 3D network has the same performance as the trained 2D network, *i.e.* after initialization the 3D network is likely to yield an inferior performance when compared to the trained 2D network. In contrast, our 2D to 3D transfer learning method does not suffer from this problem. Moreover, our 2D to 3D training strategy has a potential to be applied progressively in training a 3D model by including more and more slices, which can be done in the same spirit of what Karras *et al.* argued, *i.e.* progressively adding new layers in training GAN network for improved quality, stability, and variation [46].

In conclusion, we have made the first attempt to transfer a trained 2D CNN to a 3D counterpart for low-dose CT image denoising. The transfer learning study performed in this paper is relatively simple, but resulted in an improved and noticeable denoising performance. In the future, we plan to perform more experiments and testing with the aim of translating this initial work into clinical applications.

## Acknowledgments

The authors would like to thank the anonymous reviewers and associate editor for their constructive comments and sug-gestions, and also thank NVIDIA Corporation for the donation of Titan Xp GPU used for this research.